\documentclass[final,times,onecolumn]{elsarticle}

\usepackage[OT1]{fontenc} 
\usepackage{amsmath,bm}
\usepackage{amssymb}
\usepackage{url}

\journal{CMAME}

\begin{document}

\begin{frontmatter}



\title{Neural Network Layers for Prediction of Positive Definite Elastic Stiffness Tensors}






\author[1]{Charles F. Jekel\corref{cor1}%
    \fnref{fn1}}
\ead{jekel1@llnl.gov}

\author[1]{Kenneth E. Swartz}

\author[1]{Daniel A. White}

\author[2]{Daniel A. Tortorelli}

\author[1]{Seth E. Watts}

\cortext[cor1]{Corresponding author}
\fntext[fn1]{This document was prepared as an account of work sponsored by an agency of the United States government. Neither the United States government nor Lawrence Livermore National Security, LLC, nor any of their employees makes any warranty, expressed or implied, or assumes any legal liability or responsibility for the accuracy, completeness, or usefulness of any information, apparatus, product, or process disclosed, or represents that its use would not infringe privately owned rights. Reference herein to any specific commercial product, process, or service by trade name, trademark, manufacturer, or otherwise does not necessarily constitute or imply its endorsement, recommendation, or favoring by the United States government or Lawrence Livermore National Security, LLC. The views and opinions of authors expressed herein do not necessarily state or reflect those of the United States government or Lawrence Livermore National Security, LLC, and shall not be used for advertising or product endorsement purposes. LLNL-JRNL-832991}

\affiliation[1]{organization={Lawrence Livermore National Laboratory},
    addressline={L-227, PO Box 808},
    postcode={94551},
    city={Livermore, CA},
    country={USA}}

\affiliation[2]{organization={Department of Mechanical Sciences and Engineering},
    addressline={University of Illinois},
    city={Urbana, IL},
    postcode={61801},
    country={USA}}

\begin{abstract}
Machine learning models can be used to predict physical quantities like homogenized elasticity stiffness tensors, which must always be symmetric positive definite (SPD) based on conservation arguments. Two datasets of homogenized elasticity tensors of lattice materials are presented as examples, where it is desired to obtain models that map unit cell geometric and material parameters to their homogenized stiffness. Fitting a model to SPD data does not guarantee the model's predictions will remain SPD. Existing Cholsesky factorization and Eigendecomposition schemes are abstracted in this work as transformation layers which enforce the SPD condition. These layers can be included in many popular machine learning models to enforce SPD behavior. This work investigates the effects that different positivity functions have on the layers and how their inclusion affects model accuracy. Commonly used models are considered, including polynomials, radial basis functions, and neural networks. Ultimately it is shown that a single SPD layer improves the model's average prediction accuracy. 
\end{abstract}




\end{frontmatter}


\section{Introduction}




The use of multiscale structures has become a popular choice to generate superior designs which can now be manufactured using 3D printing techniques \cite{sanders2021optimal,LIAO2021113727,XIAO2021113949,GAO2019451}. These structures are typically designed from unit cells arrays, i.e. lattices. To optimize these structures, various surrogate models are often used to predict the unit cell's homogenized elastic response from their geometric and material properties. Most of these surrogate models do not explicitly consider whether the predicted homogenized elastic tensors are physically admissible, i.e. symmetric positive definite (SPD). This work addresses these concerns and investigates what is needed to ensure the surrogate produces SPD elasticity tensors.

Topology optimization distributes a given amount of material over a design domain to maximize a structure's performance subject to constraints (e.g. on maximum stress,  compliance). The topology optimization problem is inherently ill-posed, but restrictions or relaxation methods can be used to formulate well-posed problems.

Density based topology optimization \cite{bendsoe1989optimal,rozvany2001aims,bendsoe2013topology} use restriction whereby the material volume fraction is driven to either material or near zero in regions with a compliant ersatz material that models voids. Fine-scale oscillations in the volume fraction are prevented by imposing perimeter constraints, slope constraints, or filtration. Homogenization based methods use relaxation whereby the fine scale oscillations are used to define microstructures that vary throughout the structure \cite{NAKSHATRALA2013167}. To accommodate this the volume fraction field is replaced by a material field that allows for all possible composite materials that can be generated from mixtures of solid and ``void'' materials. As such the volume fraction varies continuously throughout the structure. This method is impractical because the optimized microstructures cannot be fabricated. However, we can make this approach practical by limiting the morphology of the microstructure, for example by using lattice microstructures.

For example \citet{WHITE20191118} designed a bridge at the macroscale level, which consisted of a lattice of Anisotruss microscale unit cells. A neural network was used to predict the unit cell's homogenized elastic response as a function of four unit cell geometric parameters. \citet{WHITE20191118} only considered a single unit-cell topology, whereas \citet{sanders2021optimal} presented a similar formulation that incorporates multiple unit cell architectures.

Various machine learning or surrogate models have been used to predict the homogenized elastic response of unit cells as functions of their geometric and material parameters \cite{WHITE20191118,watts2019simple,wang2020data,https://doi.org/10.1002/nme.6869}. For linear elastic material models, the anisotropic homogenized response reduces (in Voigt notation) to a $6\times 6$ SPD matrix. The database to produce these models is obtained from the homogenization of unit cells via highly resolved continuum finite element simulations \cite{watts2020elastic}. In a similar study \citet{watts2019simple} used polynomial based surrogate models to predict the homogenized elastic responses of the Isotruss \cite{MESSNER2016162}, the Octet truss, and the ORC truss unit cells. Neural networks have also been used to generate these models \cite{WHITE20191118}. Analogously, \citet{XIAO2021113949} used a Kriging surrogate model to predict homogenized lattice cell properties in order to design graded lattice sandwich structures. None of these studies investigate whether the surrogate model predict SPD tensors. This is important because indefinite tensors indicate unstable materials which are nonphysical for our applications. 

While our focus is on elasticity, many other physics applications require SPD matrices (e.g., heat transfer, dynamics). \citet{amsallem2009method} developed surrogate models for mass, damping, and stiffness matrices via interpolation based on an exponential manifold to preserve SPD qualities. This method guarantees SPD output by 1) taking the Eigendecomposition of a prediction, 2) applying the positive valued exponential function to the eigenvalues, and 3) reassembling the matrix with the new positive eigenvalues and the existing eigenvectors. Thus, the predicted matrices are SPD by construction. In a somewhat related approach, \citet{xu2021learning} developed SPD neural networks that also always output SPD tensors. Their approach cleverly takes advantage of the $LL^\text{T}$ Cholesky factorization, where predictions are 1) generated in lower triangular form $L$, 2) the diagonal elements of $L$ are mapped to the positive reals, and hence the 3) outputted $LL^\text{T}$ is SPD by construction. However, neither \cite{amsallem2009method} nor \cite{xu2021learning} consider whether these SPD enforcement methods sacrifice model accuracy. Additionally, they do not demonstrate what happens to model predictions if no SPD enforcement method is used.


Other methods have also been used to create SPD models. The simplest approach may be to consider linear interpolation between two SPD matrices, which can be easily proven to result in a SPD matrix. However, linear interpolation is not able to produce SPD matrices via extrapolation \cite{moakher2006symmetric}. Linear interpolation is also $C^0$ continuous, which may produce undesirable non-continuous derivatives for multi-scale topology optimization. \citet{amos2017input} presented input convex neural networks which add constraints on the network parameters to ensure that the output is a convex function. \citet{KLEIN2022104703} used input convex neural networks to model a hyperelastic material. Constraints could be constructed using SPD metrics defined in \cite{ting1996positive} to restrict neural network parameters to always output SPD matrices. However, unlike \cite{amsallem2009method} and \cite{xu2021learning}, such constrained methods do not guarantee that outputs are SPD for all possible inputs (unit cell geometries and material properties in our application) and network parameters.

The primary contribution of this work is to compare SPD enforcement methods with commonly used surrogate models \cite{queipo2005surrogate}. The specific focus only considers data points that produce SPD matrices (from specific physics applications). The paper briefly describes the creation of the SPD datasets. It is then shown that simply fitting a model to the SPD data does not produce SPD output. To remedy this, two SPD preservation methods are presented. These methods are implemented as transformation layers or functions that have no learnable parameters, hence they are easily incorporated into any surrogate model. We compare the two SPD generating surrogate models and investigate their affects on model accuracy.

\section{Elastic response of Isotruss dataset}\label{sec:models}

The elastic stiffness tensor $\mathbb{C}$ maps infinitesimal strain $\bm{\epsilon}$ to stress $\bm{\sigma}$ as $\bm{\sigma}=\mathbb{C}\bm{\epsilon}$. Using Voigt notation this mapping is expressed in matrix from as
\begin{equation}
    \begin{bmatrix}
        \sigma_{11} \\
        \sigma_{22} \\
        \sigma_{33} \\
        \sigma_{23} \\
        \sigma_{13} \\
        \sigma_{12} \\
    \end{bmatrix} = \begin{bmatrix}
        C_{11} & C_{12}        & C_{13} & C_{14} & C_{15} & C_{16} \\
               & C_{22}        & C_{23} & C_{24} & C_{25} & C_{26} \\
               &               & C_{33} & C_{34} & C_{35} & C_{36} \\
               &               &        & C_{44} & C_{45} & C_{46} \\
               & \textit{symm} &        &        & C_{55} & C_{56} \\
               &               &        &        &        & C_{66} \\
    \end{bmatrix} \begin{bmatrix}
         \epsilon_{11} \\
         \epsilon_{22} \\
         \epsilon_{33} \\
        2\epsilon_{23} \\
        2\epsilon_{13} \\
        2\epsilon_{12} \\    \end{bmatrix}.
\end{equation}
We henceforth treat $\bm{C}$ as a symmetric matrix\footnote{Symmetry ensures angular momentum is balanced.} with 21 independent components. Depending upon material symmetry, the number of independent components can be reduced. For example, an orthotropic material is simplified to
\begin{equation}\label{eq:ortho}
 \begin{bmatrix}
        C_{11} & C_{12}        & C_{13} &        &        &        \\
               & C_{22}        & C_{23} &        &        &        \\
               &               & C_{33} &        &        &        \\
               &               &        & C_{44} &        &        \\
               & \textit{symm} &        &        & C_{55} &        \\
               &               &        &        &        & C_{66} \\
    \end{bmatrix}
\end{equation}
with just 9 independent components. In all cases we assume $\bm{C}$ is SPD so that the strain energy $\bm{\epsilon}^\text{T}\bm{C}\bm{\epsilon}$ is positive \cite{ting1996positive}.

To exemplify our methodology, we develop surrogate models of the homogenized response of the orthotropic Isotruss unit-cell \cite{MESSNER2016162}. Two variants of the Isotruss are investigated, one with solid truss rods, and another with hollow rods. The constitutive materials in both unit cells are uniform and isotropic. A homogenization procedure following \cite{watts2020elastic} was used to determine the orthotropic homogenized $\bm{C}$ matrices from different rod geometries and constitutive Poisson's ratios. The homogenization procedure always produces SPD matrices. Datasets were constructed with the rod geometry and Poisson's ratio as input and the homogenized orthotropic elasticity matrix as output.

A 2D dataset was generated for solid rod Isotruss structures. The first input variable was the outer rod radius, which was sampled over $[0.001a, 0.25a)$, where $a$ is the unit cell size. Thus a 0.25 radius would have a physical length of $0.25a$. The second input was the Poisson's ratio, which was sampled over $[0, 0.5)$. One hundred random Latin hypercube samples were performed for the two variables \cite{viana2016tutorial}. The output for each pair of inputs is a $6\times 6$ SPD orthotropic homogenized elasticity matrix $\bm{C}$.

In a similar manner, a 3D dataset was generated for hollow rod Isotruss structures. The three inputs are the inner radius to outer radius ratio, the outer radius, and the Poisson's ratio. The sampling intervals were $[0.01, 0.9)$, $[0.01a, 0.25a)$, and $[0, 0.5)$ respectively. One thousand random Latin hypercube samples were performed. Like with the 2D dataset, the output is a $6\times 6$ SPD orthotropic homogenized elasticity matrix $\bm{C}$.

Models were fit to the 2D dataset and then evaluated to understand whether the models were SPD on the domain. The models learn the 9 independent components of $\bm{C}$, thus will always produce a symmetric matrix. However, it is unclear if the models will always produce SPD matrices. We implement four surrogate models:
\begin{enumerate}
    \item single layer 100 neuron model with radial basis activation functions from \cite{WHITE20191118} denoted as NN,
    \item quadratic polynomial response surface,
    \item quartic polynomial response surface,
    \item and linear kernel radial basis function (RBF) \cite{park1991universal}.
\end{enumerate}
Each of the four models were fit 1,080 times to a random fraction of 80\% of the data, with the intention to asses the probability of producing an indefinite $\bm{C}$ matrix at some portion in the domain.

\begin{figure}[hbt!]
    \centering
    \includegraphics[width=0.495\textwidth]{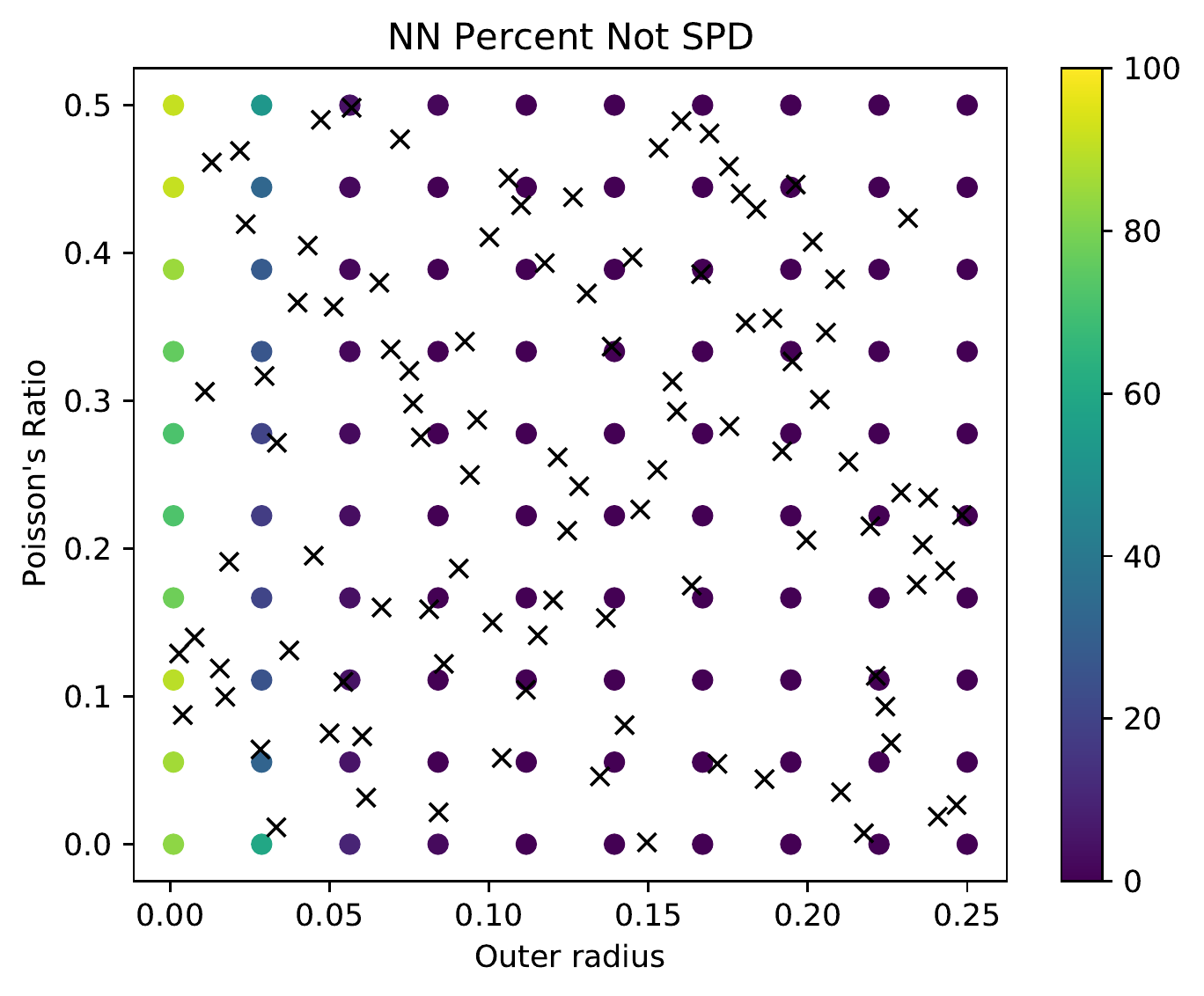}
    \includegraphics[width=0.495\textwidth]{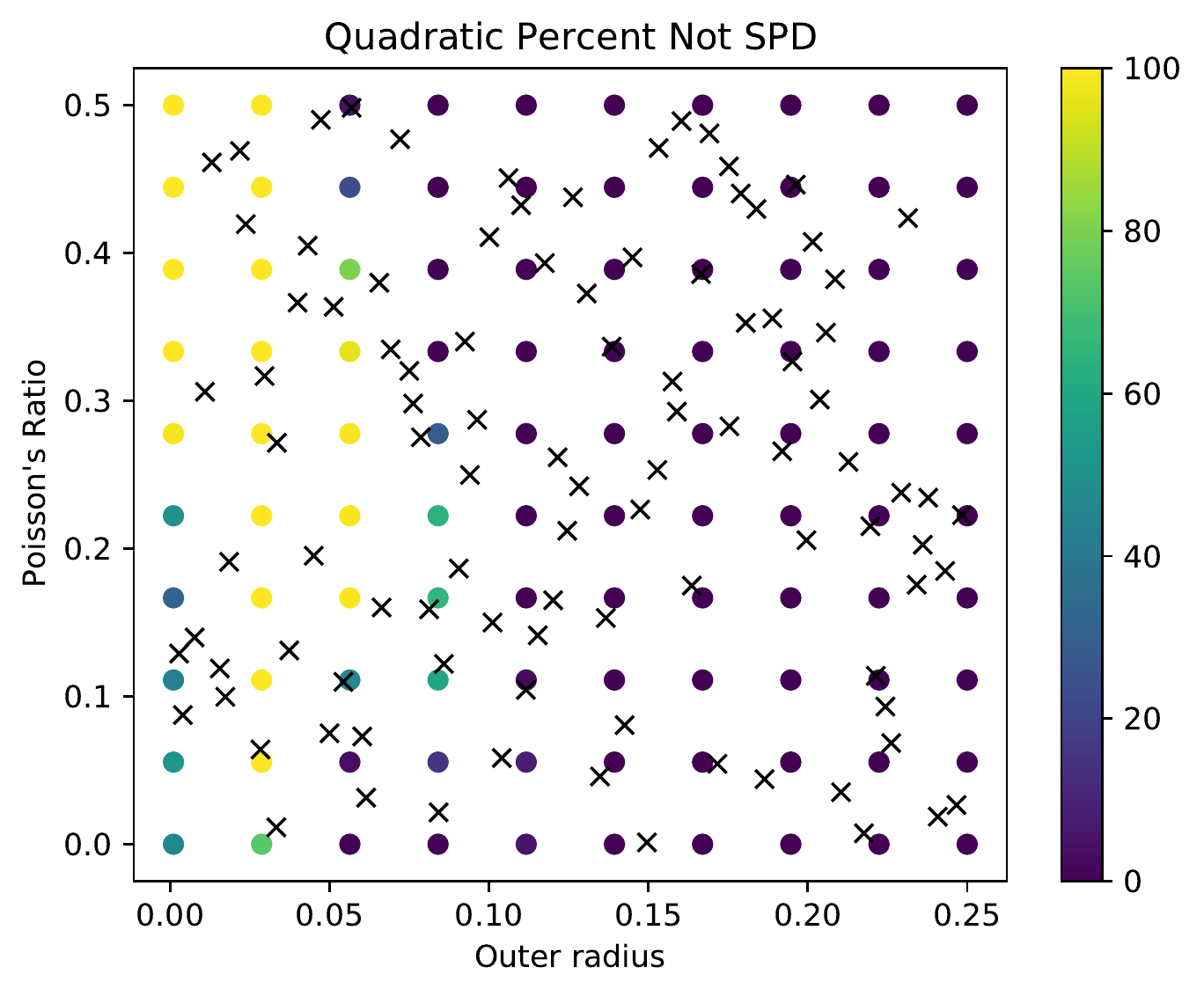}
    \includegraphics[width=0.495\textwidth]{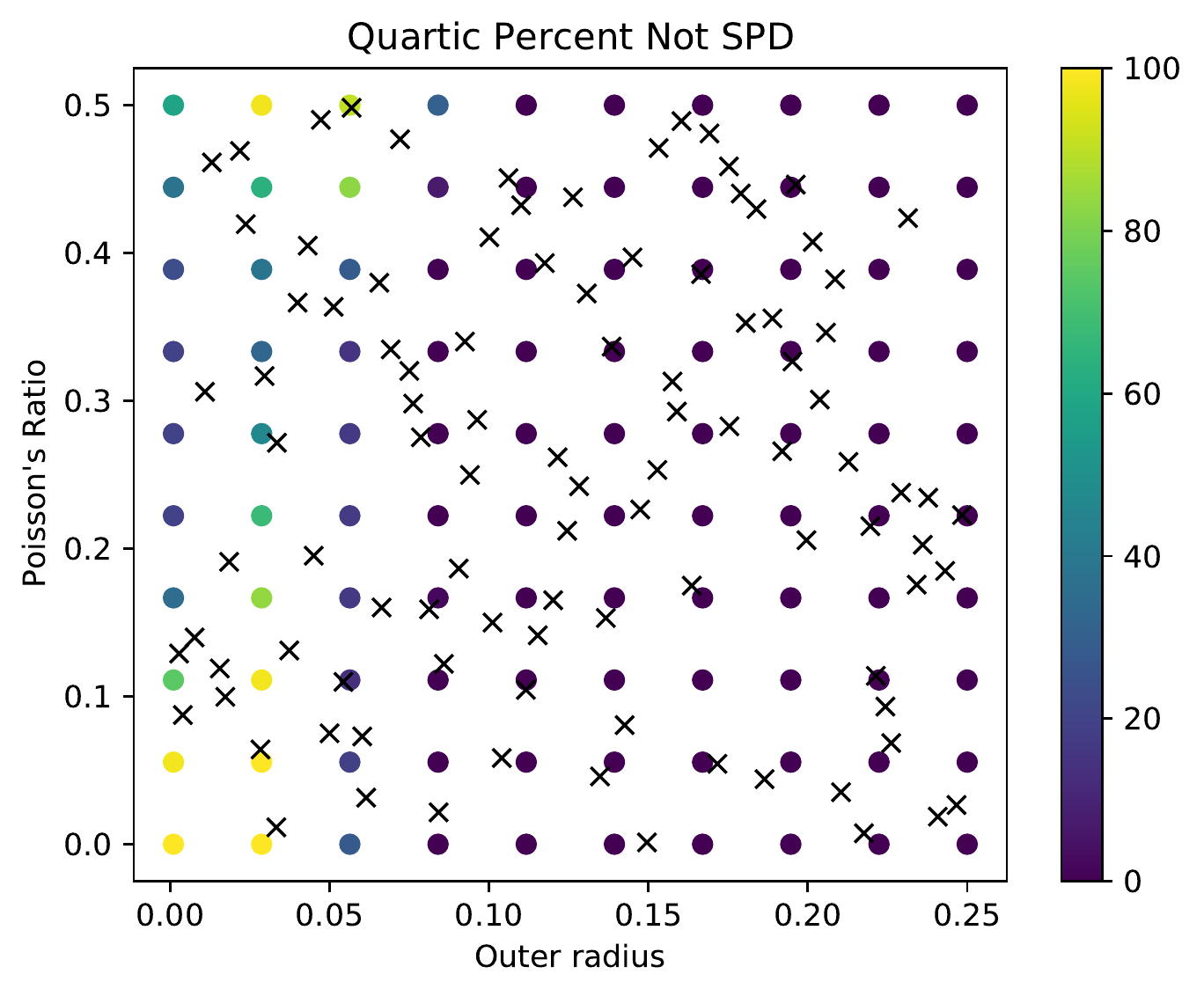}
    \includegraphics[width=0.495\textwidth]{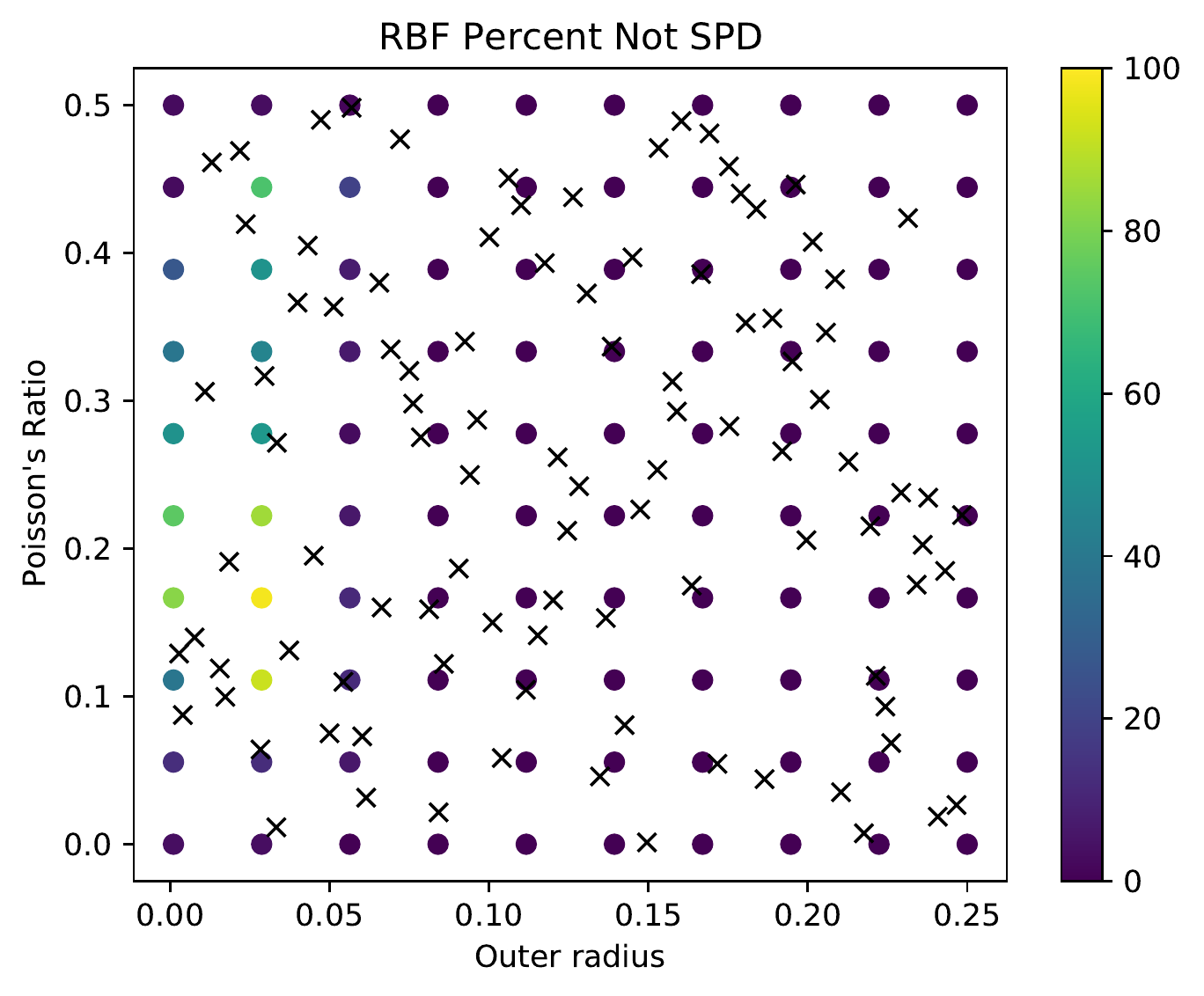}
    \caption{The percent (indicated by the dot color) of producing an indefinite $\bm{C}$  at the factorial points. The 100 data points in the 2D dataset are marked with xes.}\label{fig:nn}
\end{figure}

The trained models were then evaluated on a $10\times 10$ factorial grid indeed. All of the 1,0080 quadratic and quartic models had at least one grid point that produced a symmetric indefinite $\bm{C}$ matrix. Only one of the 1,080 neural network models and three of the 1,080 RBF models produced SPD $\bm{C}$ matrices on the entire $10\times 10$ grid. The percent of the indefinite predictions for each grid point from the 1,080 models is shown in Figure~\ref{fig:nn}. The polynomials and RBF had a single grid point that produce indefinite matrices nearly 100\% of the time. It is clear that these surrogate models do not produce SPD matrices.

\section{SPD transformations layers}

Let $\bm{\theta}$ represent the inputs to an arbitrary machine learning model, and let $\bm{\phi}$ represent the learnable parameters of such model. The goal of this work is to predict $\bm{C}(\bm{\theta}, \bm{\phi})$ where $\bm{C}$ is SPD for all $\bm{\theta}$ and $\bm{\phi}$. Borrowing the layer abstraction from neural networks, we propose SPD transformation layers to ensure SPD predictions. These layers just transform a vector input to a SPD matrix. In the following we describe the Cholesky factorization method that is based on \cite{xu2021learning}, and the Eigendecomposition that is inspired by \cite{amsallem2009method}. While \cite{amsallem2009method,xu2021learning} required transformations of the output data prior to training, models with our SPD layers can be applied to the original data without such transformations. This simplifies the error evaluation during training, and the inference in production.

Our SPD layers are mappings of a matrix (or second-order tensor) of form $\bm{X}_{bm}$ to a 3-D matrix (or third-order tensor) $\bm{C}_{bnn}$. The batch size $b$ denotes the number of samples to be transformed at a time. The relationship between number of independent matrix components $m$  and the $n \times n$ output SPD matrix dimension is expressed as the sum of the first $n$ natural numbers
\begin{equation}\label{eq:m_to_n}
    m = \frac{n(n+1)}{2}.
\end{equation}
Our work focuses on producing $6\times 6$ SPD matrices with $n=6$ and $m=21$, but these layers may also be used to produce larger SPD matrices.

\subsection{Cholesky decomposition}

The Cholesky layer uses the Cholesky decomposition to enforce SPD output. The Cholesky decomposition of a real symmetric matrix $\bm{C}$ is expressed as
\begin{equation}
    \bm{C} = \bm{L}\bm{L}^\text{T}.
\end{equation}
If the lower triangular matrix $\bm{L}$ has real values and positive diagonal values ($L_{ii} > 0$), then $\bm{L}\bm{L}^\text{T}$ produces an SPD matrix. We use this observation from \cite{xu2021learning} to construct an SPD layer that produces SPD outputs for any real input.

The Cholesky factorization layer takes in matrix $\bm{X}$ of size $*m$ and outputs a matrix $\bm{C}$ of size $*nn$, where $*$ represents an arbitrary batch size. The first step in the layer is to transform each row $\bm{X}_{*}$ of the input $\bm{X}$ to a lower triangle matrix $\bm{L}$ of shape $*nn$ as 
\begin{equation}
    \bm{X}_{*} = [x_1, x_2, \cdots, x_m] \to \bm{L}_{*} = \begin{bmatrix}
        L_{11} &        &        & \\
        L_{21} & L_{22} &        & \\ 
        \vdots & \vdots & \ddots & \\
        L_{n1} & L_{n2} & \cdots & L_{nn} \\
    \end{bmatrix}
\end{equation}where notation of $\bm{X}_{*}$ and $\bm{L}_{*}$ indicates that we are looking at a single sample from an arbitrary sized batch.

The lower triangle matrix is populated as 
\begin{equation}
    \begin{bmatrix}
        L_{11} &        &        & \\
        L_{21} & L_{22} &        & \\ 
        \vdots & \vdots & \ddots & \\
        L_{n1} & L_{n2} & \cdots & L_{nn} \\
    \end{bmatrix} = 
    \begin{bmatrix}
        p(x_1) &        &        &        & \\
        x_2    & p(x_3) &        &        & \\ 
        x_4    & x_5    & p(x_6) &        & \\
        \vdots & \vdots & \vdots & \ddots & \\
        \cdots & \cdots & \cdots & \cdots & p(x_m) \\
    \end{bmatrix}
\end{equation}
where $p : \mathbb{R} \to \mathbb{R}^+$ represents any strictly positive valued ordinary function. Such positive enforcing functions can be based on the exponential, square, ReLU, etc. Then, with matrix multiplication of the transpose, the SPD matrix is populated as
\begin{equation}
    \begin{bmatrix}
        L_{11} &        &        & \\
        L_{21} & L_{22} &        & \\ 
        \vdots & \vdots & \ddots & \\
        L_{n1} & L_{n2} & \cdots & L_{nn} \\
    \end{bmatrix} \begin{bmatrix}
        L_{11} & L_{21} & \cdots & L_{n1} \\
               & L_{22} & \cdots & L_{n2} \\ 
               &        & \ddots & \vdots \\
               &        &        & L_{nn} \\
    \end{bmatrix} = \begin{bmatrix}
            C_{11}      & C_{12} & \cdots & C_{1n} \\
                        & C_{22} & \vdots & C_{2n} \\ 
            \text{Symm} &        & \ddots & \vdots \\
                        &        &        & C_{nn} \\
        \end{bmatrix} 
\end{equation}
which is expressed in batch form as
\begin{equation}
    \bm{C}_{*} = \bm{L}_{*}\bm{L}_{*}^\text{T}.
\end{equation}
Notably $\bm{C}_{*}$ is SPD by construction.

\subsection{Eigendecomposition}
The Eigendecomposition layer enforces SPD by ensuring all predictions have positive eigenvalues. Let $\bm{A}$ be a real square symmetric matrix. If $\bm{A}$ has all positive eigenvalues, then $\bm{A}$ is SPD \cite{ting1996positive}. This principle is used to create a transformation layer that takes in real values and outputs SPD matrices.


Like the Cholesky factorization, the eigenvalue decomposition layer first takes in a size of $*m$ and outputs a size of $*nn$. The first step is to transform each row of the input into a symmetric and square matrix $\bm{A}_*$. For example the following mapping
\begin{equation}
    \bm{X}_{*} = [x_1, x_2, \cdots, x_m] \to \bm{A}_{*} = \begin{bmatrix}
        A_{11} &        &  &  &  \\
        A_{21} & A_{22} &        & \textit{symm}  & \\ 
        A_{31} & A_{23} & A_{33} &  &  \\
        \vdots & \vdots & \vdots & \ddots &  \\
        A_{n1} & A_{n2} & A_{n3} & \cdots &  A_{nn} \\
    \end{bmatrix}
\end{equation}
is obtained for for each candidate $\bm{X}_{*}$ in the batch as
\begin{equation}
    \bm{A}_{*} = \begin{bmatrix}
        x_1    &        &        &               & \\
        x_2    & x_3    &        & \textit{symm} & \\ 
        x_4    & x_5    & x_6    &               & \\
        \vdots & \vdots & \vdots & \ddots        & \\
        \cdots & \cdots & \cdots & \cdots        & x_m \\
    \end{bmatrix}.
\end{equation}
Then, an eigenvalue decomposition is performed on $\bm{A}_*$ such that
\begin{equation}
    \bm{A}_{*} = \bm{Q}_{*} \bm{\Lambda}_{*} \bm{Q}_{*}^{-1}
\end{equation}
where the columns of $\bm{Q}$ are the normalized eigenvectors and the $\Lambda_{ii}$ of the diagonal $\bm{\Lambda}$ are the corresponding eigenvalues. A positive function $p : \mathbb{R} \to \mathbb{R}^+$ (like the exponential used in \cite{amsallem2009method}) is applied to the eigenvalues
\begin{equation}
    p(\bm{\Lambda}_{*}) = \begin{bmatrix}
        p(\lambda_{11}) &                 &        & \\
                        & p(\lambda_{22}) &        & \\
                        &                 & \ddots & \\
                        &                 &        & p(\lambda_{nn}) \\
    \end{bmatrix}
\end{equation}
to ensure that each eigenvalue is strictly positive. Finally, the output matrix $\bm{C}_*$ is assembled as
\begin{equation}
    \bm{C}_{*} = \bm{Q}_{*} p(\bm{\Lambda}_{*}) \bm{Q}_{*}^{-1}
\end{equation}
which is SPD by construction.

\section{Experiments with positive enforcing functions}

This section investigates the performance of different positive enforcing functions $p : \mathbb{R} \to \mathbb{R}^+$ for a given neural network model. Specifically we use the same single layer, 100 neuron model with radial basis activation functions from \cite{WHITE20191118}, and add our SPD layer to the final prediction. Each model outputs the $6\times 6$ orthotropic matrix (with $m=9$), learning only the non-zero stiffness values in $\bm{C}$. Mean squared error was used as the loss function on the entire $6\times 6$ matrix prediction. These models were implemented in PyTorch\footnote{Our implementation of the PyTorch SPD layers as a Python package is available online at \url{https://github.com/LLNL/spdlayers}} and take advantage of automatic differentiation \cite{paszke2019pytorch}.

The positive functions $p : \mathbb{R} \to \mathbb{R}^+$ investigated are denoted in Table~\ref{table:posfuns}. A small $\epsilon=$1e-8 was added to each function to ensure positive values for all $x$. The functions are monotonically increasing for $x>0$ and satisfy either $\lim_{x\to\infty} f(x) = \lim_{x\to-\infty} f(x)$ or $\lim_{x\to-\infty} f(x)=0$. Derivatives do not exist at $x=0$ for the absolute value and ReLU functions, but do exist for the other four functions. We also note that $\lim_{x\to\infty} f(x) = x$ for the Abs, Softplus and ReLU functions.

\begin{table}[h!]
    \centering
    \begin{tabular}{c c} 
      \hline
     Name & $p(x)=$  \\
      \hline
      Abs       & $|x| + \epsilon$ \\
      Square    & $x^2 + \epsilon$ \\
      Softplus  & $\log(1 + \exp(x)) + \epsilon$ \\
      ReLU      & $\max(0, x) + \epsilon$ \\
      Quartic   & $x^4 + \epsilon$ \\
      Exp       & $\exp(x) + \epsilon$ \\
      \hline
    \end{tabular}
    \caption{Strictly positive functions used to map $x \in \mathbb{R} \to p(x) \in \mathbb{R}^+$. A small value of $\epsilon=1$e-8 was used to ensure $p(x) > 0$ for all $x\in \mathbb{R}$. Note that $\epsilon$ is not needed for the Exponential or Softplus functions, but was included in our experiment for consistency.}
    \label{table:posfuns}
\end{table}

A potentially desirable property of the ReLU function is that $p(x)=x$ when $x > 0$ resulting in minimal modification to eigenvalues through the SPD layer. Unfortunately, the ReLU function is not differentiable at $x=0$ and has a zero derivative at $x<0$ which can make training a model more difficult than other positivity functions. The function that is most similar to ReLU is the Softplus function, which also has the benefit of being continuously differentiable. The Exponential function is in a similar category as the ReLU and Softplus, but will significantly alter the magnitude of eigenvalues through the SPD layer. While the ReLU, Softplus, and Exponential function preserve the ordering of eigenvalues through the SPD layer, the remaining positivity functions may not when $x<0$. Our numerical results do not necessarily indicate this is a problem our application, however it may become problematic in other machine learning applications.

The numerical experiment repeatedly trains the same model architecture with Cholesky or Eigendecomposition to study the different positive functions. Each model positive function combination was trained with a different set of random initial weights. Additionally, for each training 20\% of the test data and 80\% of the training data were randomly assigned. The training process was repeated 1,080 times with different randomly assigned test and training data to asses their statistical performance. The results report the average ($\mu$) and standard deviation ($\sigma$) of the mean squared error computed over the testing data. Additionally, a non-parametric tolerance interval is used to report the 95\textsuperscript{th} percentile test error to 90\% confidence \cite{hong2021learning}. The conservative tolerance interval roughly represents the worst-case test error achieved by the method. This initial experiment used the (3D input) hollow truss data and all models were trained using double precision.

Two different optimization algorithms were considered. First, the deterministic gradient-based L-BFGS algorithm (implementation inspired by \cite{schmidt2005minfunc}) was used with 100 epochs. The results with L-BFGS are presented in Table~\ref{table:poslbfgs}. Additionally, the stochastic Adam algorithm described in \cite{kingma2014adam} was used with 10,000 epochs and fixed learning rates of 1e-3, 3e-4, and 1e-4. The Adam results are presented in Tables~\ref{table:postadam1e3}-\ref{table:posadam1e4}. It is obvious that the ReLU function performs the worst on this dataset, having the highest average test errors and 95\textsuperscript{th} percentiles. The other results are more difficult to interpret. 

The L-BFGS Eigendecomposition combination seemed to produce more accurate models than the L-BFGS Cholesky combination. In fact, the lowest test errors (both on average and worst-case) occurred with the L-BFGS Eigendecomposition combination with the absolute value, square, or quartic positive functions. 

\begin{table}[h!]
    \centering
    \begin{tabular}{c c c c c} 
      \hline
      & & \multicolumn{3}{c}{Test errors} \\
      SPD layer & Positivity function $p()$ & $\mu$ & $\sigma$ & $P=0.95$ \\
      \hline
     Chol. Fac. & Abs       & 0.00069 & 0.00182 & 0.00564 \\
     Chol. Fac. & Square    & 0.00135 & 0.00233 & 0.00628 \\
     Chol. Fac. & Softplus  & 0.00159 & 0.00066 & 0.00244 \\
     Chol. Fac. & ReLU      & 0.00381 & 0.00228 & 0.00753 \\
     Chol. Fac. & Quartic   & 0.00222 & 0.00265 & 0.00708 \\
     Chol. Fac. & Exp       & 0.00140 & 0.00066 & 0.00237 \\
     Eig. Decom.& Abs       & 0.00003 & 0.00005 & 0.00013 \\
     Eig. Decom.& Square    & 0.00002 & 0.00004 & 0.00008 \\
     Eig. Decom.& Softplus  & 0.00080 & 0.00035 & 0.00113 \\
     Eig. Decom.& ReLU      & 0.00202 & 0.00119 & 0.00322 \\
     Eig. Decom.& Quartic   & 0.00002 & 0.00003 & 0.00009 \\
     Eig. Decom.& Exp       & 0.00077 & 0.00052 & 0.00203 \\
      \hline
    \end{tabular}
    \caption{Comparison of positivity functions with L-BFGS to 100 training epochs.}
    \label{table:poslbfgs}
\end{table}

The Adam Cholesky combination exhibited lower errors than the L-BFGS Cholesky combination. The best observed positive functions with the former were the absolute value, Softplus, and exponential. On the other hand, only the quartic positive function performed much better with the Adam Eigendecomposition combination than the Adam Cholesky combination. Although it may not perform the best in all scenarios, Adam with a learning rate of 1e-3 produces the lowest average and worst-case errors for the data we investigated. 

\begin{table}[h!]
    \centering
    \begin{tabular}{c c c c c} 
      \hline
      & & \multicolumn{3}{c}{Test errors} \\
      SPD layer & Positivity function $p()$ & $\mu$ & $\sigma$ & $P=0.95$ \\
      \hline
     Chol. Fac. & Abs       & 0.00008 & 0.00043 & 0.00016 \\
     Chol. Fac. & Square    & 0.00049 & 0.00148 & 0.00513 \\
     Chol. Fac. & Softplus  & 0.00008 & 0.00049 & 0.00012 \\
     Chol. Fac. & ReLU      & 0.00417 & 0.00230 & 0.00795 \\
     Chol. Fac. & Quartic   & 0.00152 & 0.00240 & 0.00651 \\
     Chol. Fac. & Exp       & 0.00004 & 0.00007 & 0.00014 \\
     Eig. Decom.& Abs       & 0.00009 & 0.00008 & 0.00026 \\
     Eig. Decom.& Square    & 0.00005 & 0.00005 & 0.00017 \\
     Eig. Decom.& Softplus  & 0.00009 & 0.00026 & 0.00096 \\
     Eig. Decom.& ReLU      & 0.00491 & 0.00449 & 0.01536 \\
     Eig. Decom.& Quartic   & 0.00005 & 0.00005 & 0.00017 \\
     Eig. Decom.& Exp       & 0.00004 & 0.00012 & 0.00010 \\
      \hline
    \end{tabular}
    \caption{Comparison of positivity functions with ADAM to 10,000 training epochs and a learning rate of 1e-3.}
    \label{table:postadam1e3}
\end{table}

\begin{table}[h!]
    \centering
    \begin{tabular}{c c c c c} 
      \hline
      & & \multicolumn{3}{c}{Test errors} \\
      SPD layer & Positivity function $p()$ & $\mu$ & $\sigma$ & $P=0.95$ \\
      \hline
     Chol. Fac. & Abs       & 0.00027 & 0.00104 & 0.00307 \\
     Chol. Fac. & Square    & 0.00066 & 0.00172 & 0.00560 \\
     Chol. Fac. & Softplus  & 0.00011 & 0.00052 & 0.00022 \\
     Chol. Fac. & ReLU      & 0.00358 & 0.00232 & 0.00746 \\
     Chol. Fac. & Quartic   & 0.00128 & 0.00227 & 0.00644 \\
     Chol. Fac. & Exp       & 0.00010 & 0.00024 & 0.00045 \\
     Eig. Decom.& Abs       & 0.00012 & 0.00013 & 0.00038 \\
     Eig. Decom.& Square    & 0.00007 & 0.00007 & 0.00021 \\
     Eig. Decom.& Softplus  & 0.00016 & 0.00034 & 0.00118 \\
     Eig. Decom.& ReLU      & 0.00211 & 0.00130 & 0.00347 \\
     Eig. Decom.& Quartic   & 0.00006 & 0.00008 & 0.00020 \\
     Eig. Decom.& Exp       & 0.00008 & 0.00019 & 0.00036 \\
      \hline
    \end{tabular}
    \caption{Comparison of positivity functions with Adam to 10,000 training epochs and a learning rate of 3e-4.}
    \label{table:posadam3e4}
\end{table}

\begin{table}[h!]
    \centering
    \begin{tabular}{c c c c c} 
      \hline
      & & \multicolumn{3}{c}{Test errors} \\
      SPD layer & Positivity function $p()$ & $\mu$ & $\sigma$ & $P=0.95$ \\
      \hline
     Chol. Fac. & Abs       & 0.00066 & 0.00167 & 0.00552 \\
     Chol. Fac. & Square    & 0.00048 & 0.00137 & 0.00504 \\
     Chol. Fac. & Softplus  & 0.00014 & 0.00029 & 0.00091 \\
     Chol. Fac. & ReLU      & 0.00333 & 0.00249 & 0.00767 \\
     Chol. Fac. & Quartic   & 0.00092 & 0.00195 & 0.00626 \\
     Chol. Fac. & Exp       & 0.00029 & 0.00066 & 0.00205 \\
     Eig. Decom.& Abs       & 0.00028 & 0.00038 & 0.00127 \\
     Eig. Decom.& Square    & 0.00013 & 0.00017 & 0.00049 \\
     Eig. Decom.& Softplus  & 0.00035 & 0.00056 & 0.00158 \\
     Eig. Decom.& ReLU      & 0.00180 & 0.00075 & 0.00327 \\
     Eig. Decom.& Quartic   & 0.00013 & 0.00032 & 0.00046 \\
     Eig. Decom.& Exp       & 0.00028 & 0.00066 & 0.00196 \\
      \hline
    \end{tabular}
    \caption{Comparison of positivity functions with ADAM to 10,000 training epochs and a learning rate of 1e-4.}
    \label{table:posadam1e4}
\end{table}

\section{Experiments with SPD layers on common surrogates}

Experiments were performed to evaluate whether the inclusion of an SPD layer had detrimental effects to the model accuracy. The models from Section~\ref{sec:models} are revisited to now include the Cholesky factorization layer, Eigendecomposition layer, and again evaluated without an SPD layer. The weights are selected by minimizing the mean squared loss, using Adam with a learning rate of 1e-3 for 10,000 epochs.

Different positive functions were selected for each SPD layer. The Cholesky factorization layer used the Softplus transformation and the Eigenvalue Decomposition layer used the Square transformation function. We believe these choices produced reasonable (but not optimal) models based on our previous experiment. Indeed, the previous experiment demonstrated that the best positive function is dependent on the optimization algorithm, the dataset, and the choice of the SPD layer (i.e. Cholesky factorization or Eigendecomposition).

The surrogates are investigated on both the 2D solid truss and the 3D hollow truss datasets. The performance of the models was evaluated similarly to the previous experiment. Again the models were trained 1,080 times but here we use different training/testing ratios of 80/20, 50/50, and 10/90 to assess the models from rich to sparse data. And again, double precision was used for the data and model parameters.

\subsection{Solid truss dataset}

The results from models fit to the 2D solid truss dataset are presented in Tables~\ref{table:solid80}-\ref{table:solid10}. With one exception, we observe that the inclusion of either SPD layer improved accuracy. The quadratic polynomial with Eigendecomposition SPD layer had higher average test errors and 95\textsuperscript{th} percentiles than a quadratic polynomial with no SPD enforcement. The SPD layers had the greatest improvement of accuracy with the 10\% training and 90\% testing data split.

\begin{table}[h!]
    \centering
    \begin{tabular}{c c c c c} 
    \hline
    & & \multicolumn{3}{c}{Test errors} \\
    Model & SPD layer & $\mu$ & $\sigma$ & $P=0.95$ \\
    \hline
     NN        & Chol. Fac. & 0.00012 & 0.00038 & 0.00052 \\
     NN        & Eig. Decom.& 0.00021 & 0.00053 & 0.00111 \\
     NN        & None       & 0.00031 & 0.00086 & 0.00139 \\
     Quadratic & Chol. Fac. & 0.00142 & 0.00180 & 0.00533 \\
     Quadratic & Eig. Decom.& 0.00382 & 0.01013 & 0.02920 \\
     Quadratic & None       & 0.00326 & 0.00427 & 0.01238 \\
     Quartic   & Chol. Fac. & 0.00059 & 0.00092 & 0.00307 \\
     Quartic   & Eig. Decom.& 0.00055 & 0.00090 & 0.00296 \\
     Quartic   & None       & 0.00184 & 0.00268 & 0.00864 \\
     RBF       & Chol. Fac. & 0.00020 & 0.00054 & 0.00169 \\
     RBF       & Eig. Decom.& 0.00013 & 0.00043 & 0.00095 \\
     RBF       & None       & 0.00048 & 0.00147 & 0.00476 \\
      \hline
    \end{tabular}
    \caption{Comparison of models. 80\% Training 20\% Testing using Solid Isotruss data (100 data points, 2D).}
    \label{table:solid80}
\end{table}

\begin{table}[h!]
    \centering
    \begin{tabular}{c c c c c} 
    \hline
    & & \multicolumn{3}{c}{Test errors} \\
    Model & SPD layer & $\mu$ & $\sigma$ & $P=0.95$ \\
    \hline
     NN        & Chol. Fac. & 0.00018 & 0.00034 & 0.00090 \\
     NN        & Eig. Decom.& 0.00029 & 0.00045 & 0.00132 \\
     NN        & None       & 0.00041 & 0.00061 & 0.00202 \\
     Quadratic & Chol. Fac. & 0.00148 & 0.00082 & 0.00308 \\
     Quadratic & Eig. Decom.& 0.00353 & 0.00828 & 0.02994 \\
     Quadratic & None       & 0.00348 & 0.00185 & 0.00655 \\
     Quartic   & Chol. Fac. & 0.00072 & 0.00055 & 0.00186 \\
     Quartic   & Eig. Decom.& 0.00065 & 0.00055 & 0.00184 \\
     Quartic   & None       & 0.00207 & 0.00131 & 0.00503 \\
     RBF       & Chol. Fac. & 0.00042 & 0.00066 & 0.00202 \\
     RBF       & Eig. Decom.& 0.00027 & 0.00048 & 0.00129 \\
     RBF       & None       & 0.00074 & 0.00122 & 0.00395 \\
      \hline
    \end{tabular}
    \caption{Comparison of models. 50\% Training 50\% Testing using Solid Isotruss data (100 data points, 2D).}
    \label{table:solid50}
\end{table}

\begin{table}[h!]
    \centering
    \begin{tabular}{c c c c c} 
    \hline
    & & \multicolumn{3}{c}{Test errors} \\
    Model & SPD layer & $\mu$ & $\sigma$ & $P=0.95$ \\
    \hline
     NN        & Chol. Fac. & 0.00156 & 0.00261 & 0.00433 \\
     NN        & Eig. Decom.& 0.00193 & 0.00297 & 0.00499 \\
     NN        & None       & 0.00391 & 0.00689 & 0.01334 \\
     Quadratic & Chol. Fac. & 0.00275 & 0.00243 & 0.00717 \\
     Quadratic & Eig. Decom.& 0.00294 & 0.00288 & 0.00666 \\
     Quadratic & None       & 0.00550 & 0.00373 & 0.01174 \\
     Quartic   & Chol. Fac. & 0.00214 & 0.00197 & 0.00600 \\
     Quartic   & Eig. Decom.& 0.00213 & 0.00228 & 0.00476 \\
     Quartic   & None       & 0.00445 & 0.00394 & 0.01071 \\
     RBF       & Chol. Fac. & 0.00290 & 0.00198 & 0.00689 \\
     RBF       & Eig. Decom.& 0.00197 & 0.00177 & 0.00525 \\
     RBF       & None       & 0.00449 & 0.00382 & 0.01115 \\
      \hline
    \end{tabular}
    \caption{Comparison of models. 10\% Training 90\% Testing using Solid Isotruss data (100 data points, 2D).}
    \label{table:solid10}
\end{table}

\subsection{Hollow truss dataset}

The results from models fit to the 3D hollow truss dataset are presented in Tables~\ref{table:hollow80}-\ref{table:hollow10}. Similar to the solid truss, the results show that the inclusion of either SPD layer resulted in improved accuracy. In contrast to the solid truss, there were no observed exceptions to this trend. And surprisingly, the quadratic polynomial with Eigendecomposition SPD layer exhibited the largest accuracy improvement. The inclusion of the SPD layers with the neural network was the least impactful, as the neural network without SPD enforcement was typically more accurate than the other surrogate models with SPD layers. Regardless, the accuracy of the neural network was further improved with either SPD layer. And, we emphasize that the SPD layers ensure the desired SPD output.

\begin{table}[h!]
    \centering
    \begin{tabular}{c c c c c} 
    \hline
    & & \multicolumn{3}{c}{Test errors} \\
    Model & SPD layer & $\mu$ & $\sigma$ & $P=0.95$ \\
    \hline
     NN        & Chol. Fac. & 0.00004 & 0.00006 & 0.00013 \\
     NN        & Eig. Decom.& 0.00005 & 0.00006 & 0.00018 \\
     NN        & None       & 0.00009 & 0.00009 & 0.00029 \\
     Quadratic & Chol. Fac. & 0.00038 & 0.00020 & 0.00081 \\
     Quadratic & Eig. Decom.& 0.00044 & 0.00021 & 0.00090 \\
     Quadratic & None       & 0.00155 & 0.00059 & 0.00269 \\
     Quartic   & Chol. Fac. & 0.00012 & 0.00011 & 0.00035 \\
     Quartic   & Eig. Decom.& 0.00011 & 0.00009 & 0.00030 \\
     Quartic   & None       & 0.00045 & 0.00027 & 0.00103 \\
     RBF       & Chol. Fac. & 0.00012 & 0.00011 & 0.00036 \\
     RBF       & Eig. Decom.& 0.00009 & 0.00008 & 0.00027 \\
     RBF       & None       & 0.00023 & 0.00015 & 0.00055 \\
      \hline
    \end{tabular}
    \caption{Comparison of models. 80\% Training, 20\% Testing using Hollow Isotruss data (1,000 data points, 3D).}
    \label{table:hollow80}
\end{table}

\begin{table}[h!]
    \centering
    \begin{tabular}{c c c c c} 
    \hline
    & & \multicolumn{3}{c}{Test errors} \\
    Model & SPD layer & $\mu$ & $\sigma$ & $P=0.95$ \\
    \hline
     NN        & Chol. Fac. & 0.00004 & 0.00003 & 0.00009 \\
     NN        & Eig. Decom.& 0.00006 & 0.00003 & 0.00012 \\
     NN        & None       & 0.00009 & 0.00005 & 0.00022 \\
     Quadratic & Chol. Fac. & 0.00038 & 0.00012 & 0.00066 \\
     Quadratic & Eig. Decom.& 0.00044 & 0.00011 & 0.00064 \\
     Quadratic & None       & 0.00150 & 0.00029 & 0.00205 \\
     Quartic   & Chol. Fac. & 0.00012 & 0.00009 & 0.00038 \\
     Quartic   & Eig. Decom.& 0.00011 & 0.00005 & 0.00021 \\
     Quartic   & None       & 0.00044 & 0.00018 & 0.00083 \\
     RBF       & Chol. Fac. & 0.00015 & 0.00009 & 0.00032 \\
     RBF       & Eig. Decom.& 0.00011 & 0.00006 & 0.00023 \\
     RBF       & None       & 0.00025 & 0.00011 & 0.00046 \\
      \hline
    \end{tabular}
    \caption{Comparison of models. 50\% Training 50\% Testing using Hollow Isotruss data (1,000 data points, 3D).}
    \label{table:hollow50}
\end{table}

\begin{table}[h!]
    \centering
    \begin{tabular}{c c c c c} 
    \hline
    & & \multicolumn{3}{c}{Test errors} \\
    Model & SPD layer  & $\mu$ & $\sigma$ & $P=0.95$ \\
    \hline
     NN        & Chol. Fac. & 0.00014 & 0.00016 & 0.00039 \\
     NN        & Eig. Decom.& 0.00018 & 0.00025 & 0.00050 \\
     NN        & None       & 0.00026 & 0.00026 & 0.00068 \\
     Quadratic & Chol. Fac. & 0.00056 & 0.00021 & 0.00097 \\
     Quadratic & Eig. Decom.& 0.00070 & 0.00083 & 0.00108 \\
     Quadratic & None       & 0.00177 & 0.00024 & 0.00229 \\
     Quartic   & Chol. Fac. & 0.00025 & 0.00018 & 0.00059 \\
     Quartic   & Eig. Decom.& 0.00025 & 0.00016 & 0.00054 \\
     Quartic   & None       & 0.00069 & 0.00025 & 0.00119 \\
     RBF       & Chol. Fac. & 0.00045 & 0.00027 & 0.00102 \\
     RBF       & Eig. Decom.& 0.00036 & 0.00023 & 0.00079 \\
     RBF       & None       & 0.00064 & 0.00031 & 0.00127 \\
      \hline
    \end{tabular}
    \caption{Comparison of models. 10\% Training 90\% Testing using Hollow Isotruss data (1,000 data points, 3D).}
    \label{table:hollow10}
\end{table}


\section{Conclusion}

There are applications where the output of a prediction model is expected to be a symmetric positive definite (SPD) matrix. An example application is our prediction of homogenized elasticity tensors from orthotropic Isotruss unit cells as functions of their geometric and material parameters. We show that fitting a model to a dataset that is strictly SPD does not ensure that model predictions will be SPD. To remedy this we incorporated layers based on the Cholesky factorization and an Eigendecomposition. These methods introduce no additional learnable parameters, and ensure SPD output. The SPD layers require a strictly positive function to enforce the desired behavior. We studied several commonly used positive functions to test their efficacy. The ReLU function appeared to be the hardest to train, resulting in a lower average accuracy and larger standard deviation than the other positivity functions. Some positivity functions performed better than others depending upon the choice of optimization algorithm and machine learning architecture.

An experiment was also performed to evaluate whether the inclusion of an SPD layer affected the model accuracy. A single layer neural network, polynomials, and a linear kernel radial basis functions (RBF) were considered. The inclusion of the SPD layer increased the accuracy for any particular surrogate model.

\section*{Acknowledgment}
This work was performed under the auspices of the U.S. Department of Energy by Lawrence Livermore National Laboratory under Contract DE-AC52-07NA27344 and was supported by the LLNL-LDRD Program under Project No. 20-ERD-020.



 \bibliographystyle{elsarticle-num-names} 
 \bibliography{spd}





\end{document}